\documentclass{llncs}

\usepackage{epsfig}

\newcommand{\Natural}{\mathop{\rm I\kern-.2emN}}
\newcommand{\Real}{\mathop{\rm I\kern-.2emR}}

\newcommand{\Sol}[0]{{\cal S}} 
\newcommand{\Vois}[0]{{\cal V}} 

\begin{document}

\title{Neutral Fitness Landscape in the Cellular Automata Majority Problem}
\titlerunning{Neutrality in Majority Problem}
\author{S. Verel$^{1}$ \and P. Collard$^{1}$ \and M. Tomassini$^{2}$ \and L. Vanneschi$^{3}$}
\authorrunning{} 
\institute{Universit\'{e} de Nice-Sophia Antipolis/CNRS \email{(verel,pc)@i3s.unice.fr}
\and University of Lausanne \email{Marco.Tomassini@unil.ch} \and University of Milano \email{vanneschi@disco.unimib.it}}

\maketitle

\begin{abstract}
We study in detail the fitness landscape of a difficult cellular automata
computational task: the majority problem. Our results show why this
problem landscape is so hard to search, and we quantify the large degree
of neutrality found in various ways. We show that a particular
subspace of the solution space, called the "Olympus", is where good solutions concentrate,
and give measures to quantitatively characterize this subspace.

\end{abstract}

\section{Introduction} 
\label{intro}

Cellular automata (CAs) are discrete dynamical systems that have been studied
for years due to their architectural simplicity and the wide spectrum
of behaviors they are capable of \cite{wolfram-book-02}.
Here we study CAs that can be said to perform a simple ``computational'' task. 
One such task is the so-called \textit{majority} or \textit{density} task in which a
two-state CA is to decide
whether the initial state contains more zeros than ones or vice versa. In spite of
its apparent simplicity, it is a difficult problem for a CA as
it requires a coordination among the automata. As such, it is a perfect paradigm
of the phenomenon of \textit{emergence} in complex systems. That is, the task solution is
an emergent global property of a system of locally interacting agents.
Indeed, it has been proved that no CA can perform the task perfectly i.e., for 
any possible initial binary
configuration of states \cite{landbelew95}. However, several efficient CAs for the density task
have been found either by hand or by using heuristic methods, especially evolutionary
computation \cite{mitchelletal94a,andreetal96b,juille98}.
For a recent review see \cite{crutch-mitch-das-03}.

All previous investigations have empirically shown that finding good CAs
for the majority task is very hard. 
However, there have been no
investigations, to our knowledge, of the reasons that make this particular fitness
landscape a difficult one. In this paper we statistically quantify in
various ways the degree of difficulty of searching the majority CA landscape. 

The paper proceeds as follows. 
The next section summarizes some known
facts about CAs for the density task. A description of its fitness landscape follows,
focusing on the hardness and neutrality aspects.
Next we identify 
and analyze a particular subspace of the problem search space 
called the Olympus. 
Finally, we present our conclusions and hints to further works and open questions.

\section{The Majority Problem} 
\label{maj}

The density task is a prototypical distributed
computational problem for CAs. For a finite CA of size $N$ it is defined as follows.
Let $\rho_0$ be the fraction of 1s in the \textit{initial configuration} (IC) ${\bf s}_{0}$.
The task is to determine whether $\rho_0$ is greater than or less than $1/2$. In
this version, the problem is also known as the \textit{majority} problem.
If $\rho_0 > 1/2$ then the CA must
relax to a fixed-point configuration of all 1's that we indicate as
$(1)^N$; otherwise it must relax to a fixed-point
configuration of
all 0's, noted $(0)^N$, after a number of time steps of the order of the grid size
$N$. Here $N$ is set to 149, the value that has been customarily used in research on the
density task (if $N$ is odd one avoids the case $\rho_0=0.5$ for which the problem is undefined).

This computation is trivial  for a computer having a central
control. Indeed, just scanning the array and adding up the number
of, say, 1 bits will provide the answer in $O(N)$ time. However, it
is nontrivial for a small radius one-dimensional CA since such a CA can only transfer
information at finite speed relying on local information exclusively, while density
is a global property of the configuration of states.
It has been shown
that the density task cannot be solved perfectly by a uniform, two-state CA
with finite radius \cite{landbelew95}.

The lack of a perfect solution does not prevent one from searching for imperfect solutions
of as good a quality as possible. In general, given a desired global behavior for a
CA (e.g., the density task), it is extremely difficult to infer the local
CA rule that will give rise to the emergence of the computation sought. This is
because of the possible nonlinearities and large-scale collective effects that
cannot in general be predicted from the sole local CA updating rule, even if it is
deterministic. Since exhaustive evaluation of all possible rules is out of the
question except for elementary ($d=1,r=1$) and perhaps radius-two automata, one possible solution  consists in using evolutionary algorithms, as first proposed by Packard in
\cite{packard88} and further developed by Mitchell et al. 
\cite{mitchelletal94a,crutch-mitch-das-03}.\\
The \textit{standard performance} of the best rules (with $r=3$) found at the end of the evolution is defined as the fraction of correct classifications over $n=10^4$ randomly chosen ICs. 
The ICs are sampled
according to a binomial distribution (i.e., each bit is independently drawn with
probability $1/2$ of being 0). \\
Mitchell and coworkers performed a number of studies on the emergence of
synchronous CA strategies for the density task (with $N=149$) during evolution
\cite{crutch-mitch-das-03,mitchelletal94a}.
Their results are significant since they represent one of the few instances where the
dynamics of emergent computation in complex, spatially extended systems can be
understood.
As for the evolved CAs, it was noted that, in most runs, the GA found
unsophisticated strategies that consisted in expanding sufficiently large blocks
of adjacent 1s or 0s. This ``block-expanding'' strategy is unsophisticated in that
it mainly uses local information to reach a conclusion. As a consequence, only those
IC that have low or high density are classified correctly since they are more likely
to have extended blocks of 1s or 0s. These CAs have a performance around
$0.6$. A few runs yielded more
sophisticated  CAs with performance (around $0.77$) on a wide
distribution of ICs. However, high-performance
automata have evolved only nine times out of $300$ runs of the genetic algorithm. 
This clearly shows that the search space is a very difficult one,
even there exists some recent works on coevolutionary algorithm \cite{pagie01comparison} 
which able to find a number of ``block expanding'' strategies.

These sophisticated strategies rely on traveling signals (``particles'') that transfer
spatial and temporal information about the density in local regions through the
lattice, and have been quantitatively described with a framework known as ``computational mechanics'' \cite{dasetal94,hansonCrutch95}.
The GKL rule \cite{gacsEtal78}
is hand-coded  but its behavior is similar to that of the best solutions found
by evolution. Das and Davis solutions are two other good solutions that have been found by hand
\cite{crutch-mitch-das-03}. 
Other researchers have been able to artificially
evolve a better CA ($ABK$) by using genetic programming\cite{andreetal96b}. 
Finally, Juill\'e \textit{et al} \cite{juille98} obtained
still better CAs ($Coe1$ and $Coe2$) by using a coevolutionary algorithm. Their coevolved CA has performance
about 0.86, which is the best result known to date. We call the six best local optima known, with a standard performance
over $0.81$, the \textit{blok} (tab. \ref{tab-opt6}).

In the next section we present a study
of the overall fitness landscape, while section \ref{olympus} concentrates on the structure of the landscape around the \textit{blok}.

\begin{table}
\caption{Description in hexadecimal 
and standard performance of the 6 previously known best rules (\textit{blok}) computed on sample size of $10^4$.\label{tab-opt6}}

\begin{center}
\begin{tabular}{|c|c|}
\hline
GKL 0.815 & Das  0.823 \\

{\scriptsize 
005F005F005F005F005FFF5F005FFF5F}
&
{\scriptsize
009F038F001FBF1F002FFB5F001FFF1F}\\

\hline
Davis 0.818 & ABK  0.824 \\
{\scriptsize
070007FF0F000FFF0F0007FF0F310FFF}
&
{\scriptsize
050055050500550555FF55FF55FF55FF}\\

\hline
Coe1  0.851 & Coe2  0.860 \\
{\scriptsize
011430D7110F395705B4FF17F13DF957}
&
{\scriptsize
1451305C0050CE5F1711FF5F0F53CF5F}\\
\hline
\end{tabular}

\end{center}
\end{table}

\section{Fitness Landscape and Neutrality of the Majority Task}

First we recall a few fundamental concepts about fitness landscapes \cite{reidys01neutrality}.
A {\it fitness landscape} is a triplet $(\Sol, \Vois, f)$ such that :
$\Sol$ is the set of potential solutions,
$\Vois : \Sol \rightarrow 2^\Sol$ is the neighborhood function 
which associates to each solution $s \in \Sol$ a set of neighbor solutions $\Vois(s) \subset S$,
$f : \Sol \rightarrow \Real$ is the fitness function 
which associates a real number to each solution.

Within the framework of metaheuristic by local search, 
the local operators allow to define the neighborhood $\Vois$.
If the metaheuristic only uses one operator $op$,
the neighborhood of a solution $s$ is often defined as $\Vois(s) = \{ s^{'} \in \Sol ~|~ s^{'} = op(s) \}$.
If more than one operator are used,
it is possible to associate one fitness landscape to each operator
or to define the set of neighbors as the set of solutions obtained by one of the operators.
A neighborhood could be associated to a distance;
for example,
in the field of genetic algorithms,
when the search space is the set of bit strings of fixed size,
the operator which change the value of one bit defines the neighborhood.
Thus,
two solutions are neighbors if their Hamming distance is equal to 1.

The notion of neutrality has been suggested by Kimura \cite{KIM:83} in his study of the evolution 
of molecular species. According to this view, most mutations are either neutral 
(their effect on fitness is small) or lethal.
In the analysis of fitness landscapes, the notion of neutral mutation appears to be useful
\cite{reidys01neutrality}. Let us thus define more precisely the notion of neutrality for
fitness landscapes.

A \textit{test of neutrality} is a predicate $isNeutral : 
S \times S \rightarrow \lbrace true, false  \rbrace$ that assigns to every $(s_1, s_2) \in S^2$ 
the value $true$ if there is a small difference between $f(s_1)$ and $f(s_2)$.

For example, usually $isNeutral(s_1, s_2)$ is $true$ if $f(s_1) = f(s_2)$. 
In that case, $isNeutral$ is an equivalence relation. 
Other useful cases are $isNeutral(s_1, s_2)$ is $true$
 if $|f(s_1) - f(s_2)| \leq 1/M$ with $M$ is the population size.
When $f$ is stochastic, $isNeutral(s_1, s_2)$ is $true$ if $|f(s_1) - f(s_2)|$ is under the evaluation error.

For every $s \in S$, the \textit{neutral neighborhood} of $s$ is the set 
$\Vois_{neut}(s) = \lbrace s^{'} \in \Vois(s) ~|~ isNeutral(s,s^{'}) \rbrace$ and the \textit{neutral degree}
 of $s$, noted $nDeg(s)$ is the number of neutral neighbors of $s$, 
 $nDeg(s) = \sharp(\Vois_{neut}(s) - \lbrace s \rbrace$).\\
 A fitness landscape is neutral if there are many solutions with high neutral degree.
In this case, we can imagine fitness landscapes with some plateaus called 
\textit{neutral networks}. There is no significant difference of fitness between solutions on neutral networks and the population drifts around on them.

A \textit{neutral walk} $W_{neut} = (s_0, s_1, \ldots, s_m)$
is a walk where 
for all $ i \in [0, m-1]$, $s_{i+1} \in \Vois(s_i)$ and
for all $(i,j) \in [ 0, m ]^2$ , $isNeutral(s_i, s_j)$ is $true$.\\
A \textit{Neutral Network}, denoted $NN$, is a graph $G=(V,E)$ where the set $V$ of vertices is 
the set of solutions belonging to $S$ such that for all $s$ and $s^{'}$ from $V$ there is a neutral
walk $W_{neut}$ belonging to $V$ from $s$ to $s^{'}$, and two vertices are connected by an edge of $E$ if 
they are neutral neighbors.

\subsection{Statistical Measures of neutrality}
\label{measures}

H. Ros\'e et al. \cite{rose96} develop the \textit{density of states} approach
(DOS) by plotting the number of sampled solutions in the search space with the same fitness 
value. 
Knowledge of this density allows to evaluate the performance of random search or random initialization
of metaheuristics. DOS gives the probability of having a given fitness value when a
solution is randomly chosen. The tail of the distribution at optimal fitness value gives
a measure of the difficulty of an optimization problem: the faster the decay, the harder the problem.

To study the neutrality of fitness landscapes, we should be able to measure and describe
a few properties of $NN$. The following quantities are useful. The \textit{size} $\sharp NN$ i.e., the
number of vertices in a $NN$,
the \textit{diameter}, which is the maximum distance between two solutions belonging to $NN$. 
The \textit{neutral degree distribution} of solutions
is the degree distribution of the vertices in a $NN$. Together with the size and the diameter, it
gives information which plays a
role in the dynamics of metaheuristic \cite{NIM:99}.
Another way to describe $NN$ is given by the \textit{autocorrelation of neutral degree} along a
neutral random walk \cite{bastolla03}. 
At each step $s_i$ of the walk,
one neutral solution $s_{i+1} \in \Vois(s_i)$ is randomly chosen 
such as $\forall j \leq i$, $isNeutral(s_j, s_j)$ is true.
From neutral degree collected along this neutral walk, 
we computed its autocorrelation. 
The autocorrelation  measures the correlation structure of a $NN$.
If the correlation is low, the variation of neutral degree is low ;
and so, there is some areas in $NN$ of solutions which have nearby neutral degrees. \\

\section{Neutrality in the Majority Problem landscape} 

In this work we use a performance measure, the \textit{standard performance} defined in
section \ref{maj}, which is based on the fraction of $n$ initial
configurations that are correctly classified from one sample.
Standard performance is a hard measure
because of the predominance in the sample of ICs close to $0.5$ and
it has been typically employed to measure a CA's capability on the density task.

The error of evaluation leads us to define the neutrality of the landscape. 
The standard performance cannot be known perfectly due to random variation of samples of ICs. 
The ICs are chosen independently, so the fitness value $f$ of a solution follows
a normal law $\mathcal{N}(f, \frac{\sigma(f)}{\sqrt{n}})$, where $\sigma$ is the standard deviation
 of sample of fitness $f$, and $n$ is the sample size. 
For binomial sample, $\sigma^2(f) = f (1-f)$, the variance of Bernouilli trial.
Thus two neighbors $s$ and $s^{'}$ are neutral neighbors ($isNeutral(s,s^{'})$ is $true$) if a t-test accepts
 the hypothesis of equality of $f(s)$ and $f(s^{'})$ with $95$ percent of confidence.
 The maximum number of fitness values statistically different for standard performance
 is $113$ for $n=10^4$, $36$ for $n=10^3$ and $12$ for $n=10^2$.\\

\subsection{Analysis of the Full Landscape}
\label{full}

\paragraph{Density Of States.}

It has proved difficult to obtain information on the Majority Problem landscape by random sampling
due to the large number of solutions with zero fitness.
From $4.10^3$ solutions using the uniform random sampling technique,
$3979$ solutions have a fitness value equal to 0. 
Clearly, the space appears to be a difficult one to search since the tail of the distribution to the right is non-existent.
Figure \ref{fig-neutr}-a shows the DOS obtained using the Metropolis-Hastings technique for importance sampling. 
For the details of the techniques used to sample high fitness values of the space, see \cite{gecco-04}.
This time, over the $4.10^3$ solutions sampled, only $176$
have a fitness equal to zero, and the DOS clearly shows a more
uniform distribution of rules over many different fitness values.
It is important to remark a considerable number of solutions
sampled with a fitness approximately equal to $0.5$.
Furthermore, no solution with a fitness value superior to $0.55$
has been sampled.

Computational costs do not allow us to analyse many neutral networks.
In this section we analyse two important large neutral networks ($NN$). 
A large number of CAs solve the majority density problem on only half of ICs because
they converge nearly always on the final configuration $(O)^N$ or $(1)^N$ and thus have performance
about $0.5$. Mitchell et al. \cite{mitchelletal94a}
call these ``default strategies'' and notice that they are the first stage in the evolution of the
population before jumping to higher performance values associated to ``block-expanding'' strategies
(see section \ref{maj}). We will study this large $NN$, denoted $NN_{0.5}$ around standard
performance $0.5$ to
understand the link between $NN$ properties and GA evolution. The other $NN$, denoted $NN_{0.76}$,
is the $NN$ around fitness $0.7645$ which contains one neighbor of a CA found by Mitchell et al. 
The description of this ``high'' $NN$ could give clues as how to ``escape'' from $NN$ toward even
higher fitness values.

\paragraph{Diameter.}

In our experiments, we perform $5$ neutral walks on $NN_{0.5}$ and $19$ on $NN_{0.76}$. Each neutral
walk has the same starting point on each $NN$. We try to explore the $NN$ by strictly increasing the Hamming
distance from the starting solution at each step of the walk. The neutral walk stops when there is no
neutral step that increases distance. The maximum length of walk is thus $128$.
On average, the length of neutral walks on $NN_{0.5}$ is $108.2$ and $33.1$ on $NN_{0.76}$. 
The diameter of $NN_{0.5}$ is thus larger than the one of $NN_{0.76}$.

\begin{figure}[!ht]
\begin{center}
\begin{tabular}{ccc}
\mbox{
  \epsfig{figure=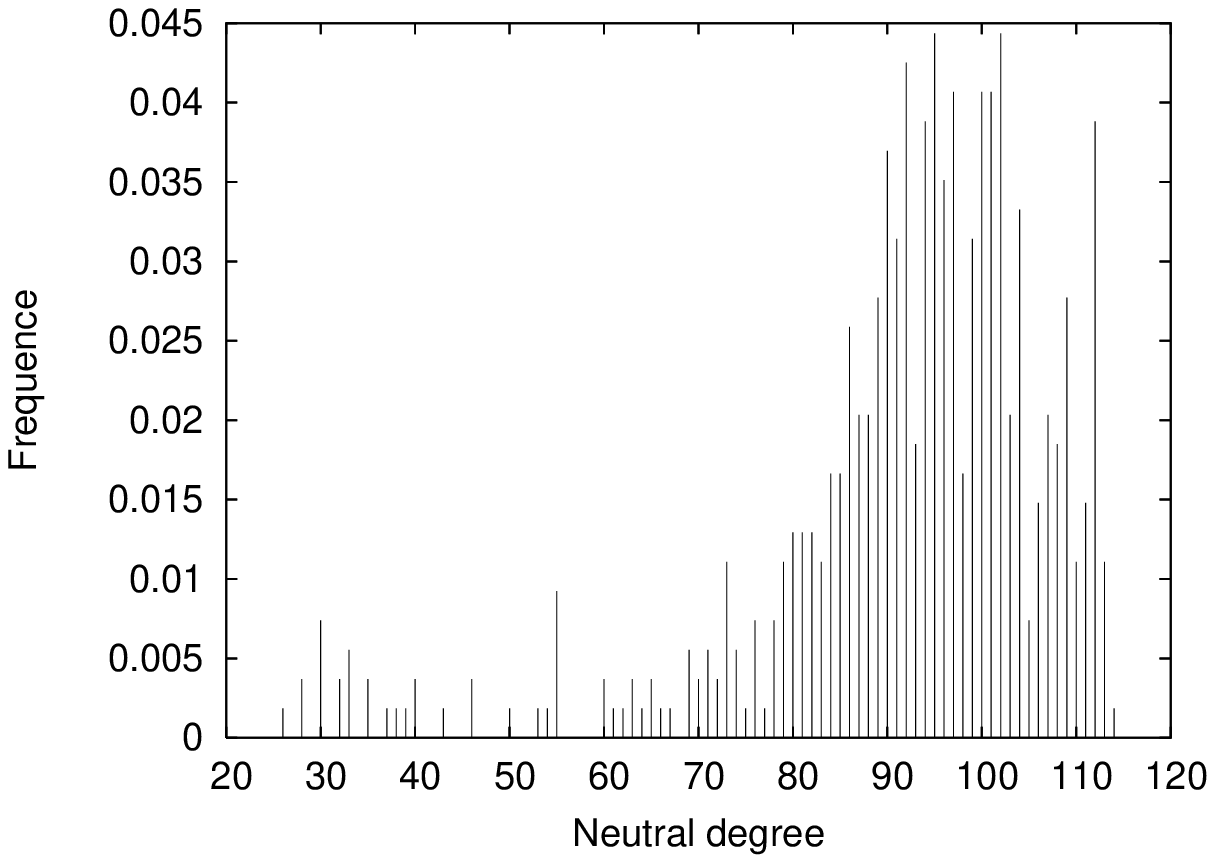,width=5.6cm,height=5.5cm} } & &
\mbox{
  \epsfig{figure=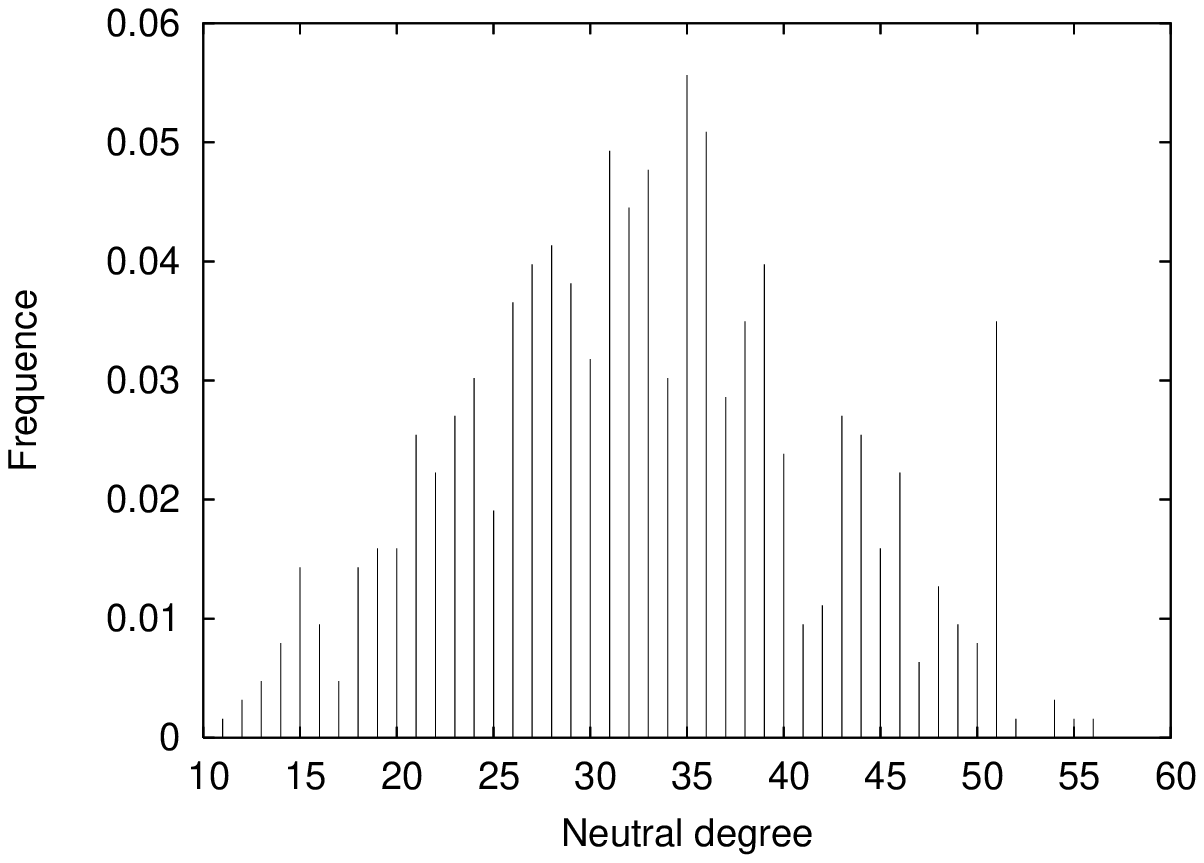,width=5.6cm,height=5.5cm} }  \protect  \\
(a)   & &  (b)   \\
\end{tabular}
\end{center}
\caption{Distribution of Neutral Degree along all neutral walks on $NN_{0.5}$ in (a) and $NN_{0.76}$ in (b).
\label{fig-nDeg_distri} }
\end{figure}

\paragraph{Neutral Degree Distribution.}

Figure \ref{fig-nDeg_distri} shows the distribution of neutral degree collected along all neutral walks.
The distribution is close to normal for $NN_{0.76}$. For $NN_{0.5}$ the distribution
is skewed and approximately bimodal with a strong peak around $100$ and a small peak around $32$. 
The average of neutral degree on
$NN_{0.5}$ is $91.6$ and standard deviation is $16.6$; on $NN_{0.76}$, the average is $32.7$ and 
the standard deviation is $9.2$.
The neutral degree for $NN_{0.5}$ is very high : $71.6$ \% of neighbors are neutral neighbors.
For $NN_{0.76}$, there is $25.5$ \% of neutral neighbors. It can be compared to the average 
neutral degree of the neutral $NKq$-landscape with $N=64$, $K=2$ and $q=2$ which is $33.3$ \% . 

\begin{figure}[!ht]
\begin{center}
\begin{tabular}{ccc}
\mbox{
  \epsfig{figure=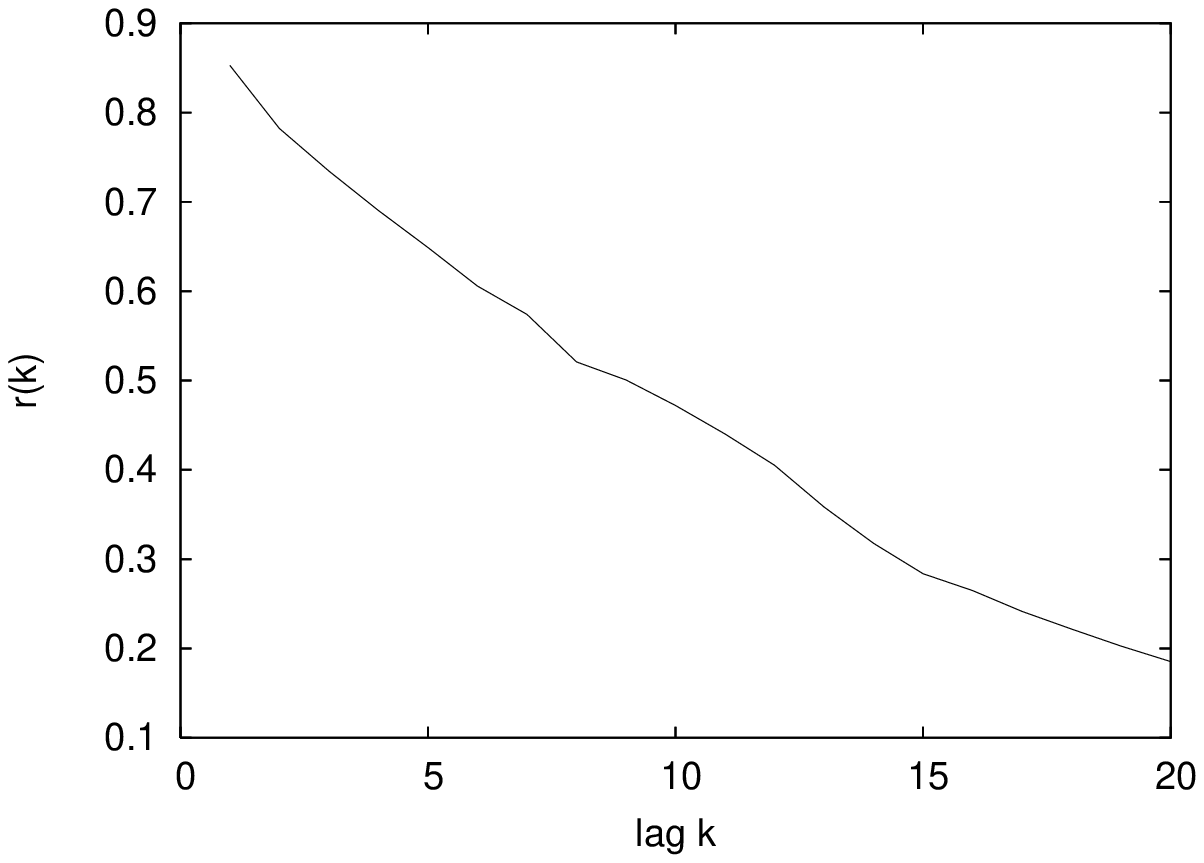,width=5.6cm,height=5.5cm} } & &
\mbox{
  \epsfig{figure=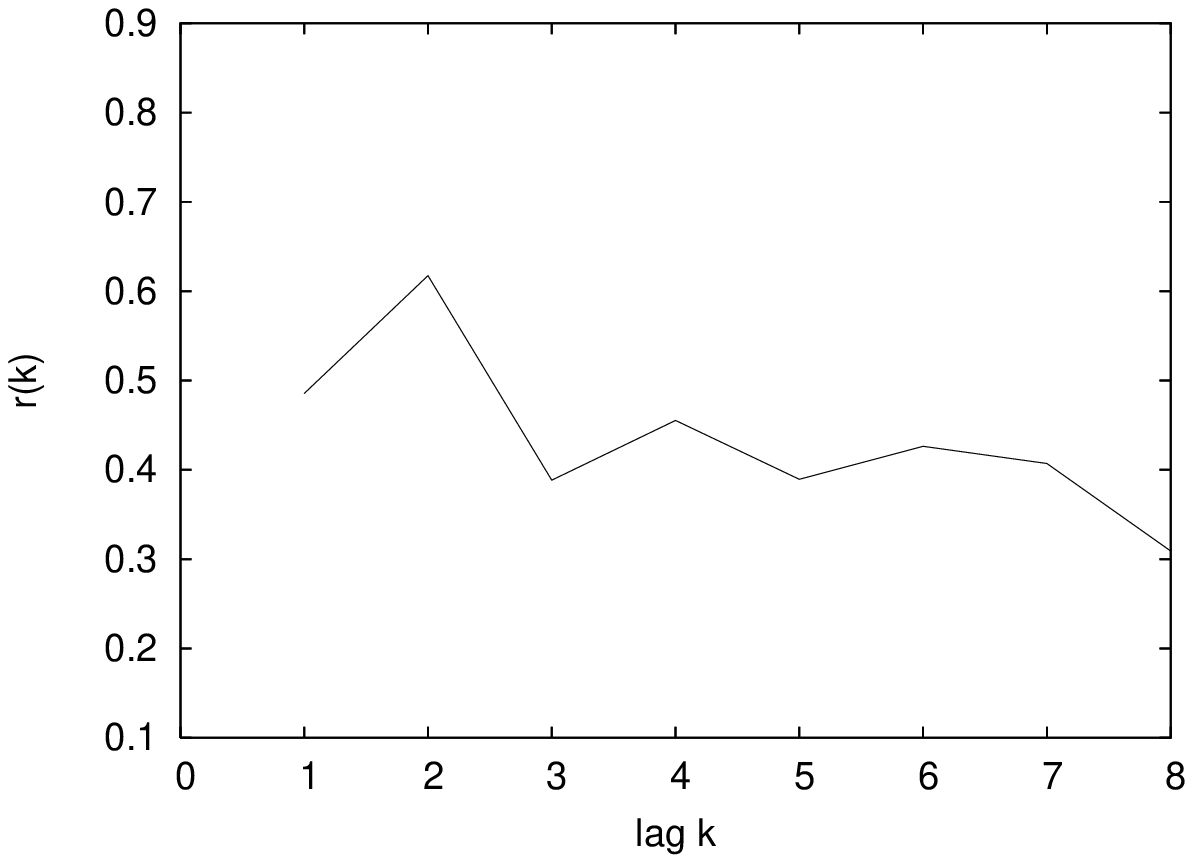,width=5.6cm,height=5.5cm} }  \protect  \\
(a)   & &  (b)   \\
\end{tabular}
\end{center}
\caption{Estimation of the autocorrelation function of neutral degrees along neutral random walks for 
$NN_{0.5}$ (a) and for $NN_{0.76}$ (b).
\label{fig-nDeg_acf} }
\end{figure}

\paragraph{Autocorrelation of Neutral Degree.}

Figure \ref{fig-nDeg_acf} gives an estimation of the autocorrelation function $\rho(k)$  of neutral
degree of the neutral networks. The autocorrelation function is computed for each neutral walk 
and the estimation $r(k)$ of $\rho(k)$ is given by the average of $r_i(k)$ over all autocorrelation functions. 
For both $NN$, there is correlation. The correlation is higher for $NN_{0.5}$ ($r(1)=0.85$) than  for $NN_{0.76}$ ($r(1)=0.49$).
From the autocorrelation of the neutral degree, one can conclude that the neutral network
topology is not completely random, since otherwise correlation should have been nearly equal to zero. 
Moreover, the variation of neutral degree is smooth on $NN$; in other words, the neighbors in $NN$ have nearby neutral degrees. So, there is some area where the neutral degree is homogeneous.

This study give us a better description of Majority fitness landscape neutrality which have important
consequence on metaheuristic design.
The neutral degree is high. Therefore, the selection operator should take into account the case of equality of
fitness values. Likewise the mutation rate and population size should fit to this neutral degree in order to
find rare good solutions outside $NN$ \cite{barnett01netcrawling}.
For two potential solutions $x$ and $y$ on $NN$, the probability $p$ that at least one solution escaped
from $NN$ is $P(x \not\in NN \cup y \not\in NN) = P(x \not\in NN) + P(y \not\in NN) - P(x \not\in NN \cap y \not\in NN)$. 
This probability is higher when solutions $x$ and $y$ are far due to the correlation of neutral degree in $NN$. 
To maximize the probability of escaping $NN$ the distance between potential solutions of population should be as far as possible on $NN$.
The population of an evolutionary algorithm should spread over $NN$.

\subsection{Study on the Olympus Landscape}
\label{olympus}

In this section we show that there are many similarities inside the \textit{blok} (see section
\ref{maj}), and we use
this feature to define what we have named the \textit{Olympus Landscape}, a subspace of
the full landscape in which good solutions are found. Next, we study
the relevant properties of this subspace.
Before defining the Olympus we study the two natural symmetries of the majority problem.

The states 0 and 1 play the same role in the computational task; so flipping bits in the entry of a rule and in the result have no effect on performance. In the same way, CAs can compute the majority problem according to right or left direction without changing performance. 
We denote $S_{01}$ and $S_{rl}$ respectively the corresponding operator of $0/1$ symmetry and $right/left$ symmetry. Let $x=(x_0,\ldots,x_{\lambda-1}) \in \lbrace 0,1 \rbrace^{\lambda}$ be a solution with $\lambda=2^{2r+1}$.
The $0/1$ symmetric of $x$ is $S_{01}(x) = y$ where for all $i$, $y_i = 1 - x_{\lambda-i}$.
The $right/left$ symmetric of $x$ is $S_{rl}(x) = y$ where for all $i$, $y_i = x_{\sigma(i)}$ 
with $\sigma(\sum_{j=0}^{\lambda-1} 2^{n_j}) = \sum_{j=0}^{\lambda-1} 2^{\lambda-1 - n_j}$.
The operators are commutative: $S_{rl} S_{01} = S_{01} S_{rl}$. From the $128$ bits, $16$ are invariant by $S_{rl}$ and none by $S_{01}$. 

Two optima from the \textit{blok} could be distant whereas some of theirs symmetrics are closer. 
Here the idea is to choose for each \textit{blok} one symmetric in order to broadly maximize the number of joint bits.

The optima GKL, Das, Davis and ABK have $2$ symmetrics only because symmetrics by $S_{01}$ and $S_{rl}$ are equal. The optima Coe1 and Coe2 have $4$ symmetrics. 
So, there are $2^4.4^2=256$ possible sets of symmetrics. 
Among these sets, 
we establish the maximum number of joint bits which is possible to obtain is $51$. 
This ``optimal'' set contains the six \textit{Symmetrics of Best Local Optima Known} (\textit{blok$^{'}$}) which are
GKL$^{'}$ = GKL, 
Das$^{'}$ = Das,
Davis$^{'}$ = $S_{01}($Davis$)$,
ABK$^{'}$ = $S_{01}($ABK$)$,
Coe1$^{'}$ = Coe1 and
Coe2$^{'}$ = $S_{rl}($Coe2$)$.

The Olympus Landscape is defined from the \textit{blok$^{'}$} as the subspace of dimension $77$ defined by the string $S^{'}$:
\begin{center}
{ \scriptsize
000*0*0* 0****1** 0***00** **0**1** 000***** 0*0**1** ******** 0*0**1*1\\
0*0***** *****1** 111111** **0**111 ******** 0**1*1*1 11111**1 0*01*111}\\
\end{center}

\paragraph{Density Of States.}

\begin{figure}[!ht]
\begin{center}
\begin{tabular}{cc}
\includegraphics[height=5cm,width=6cm]{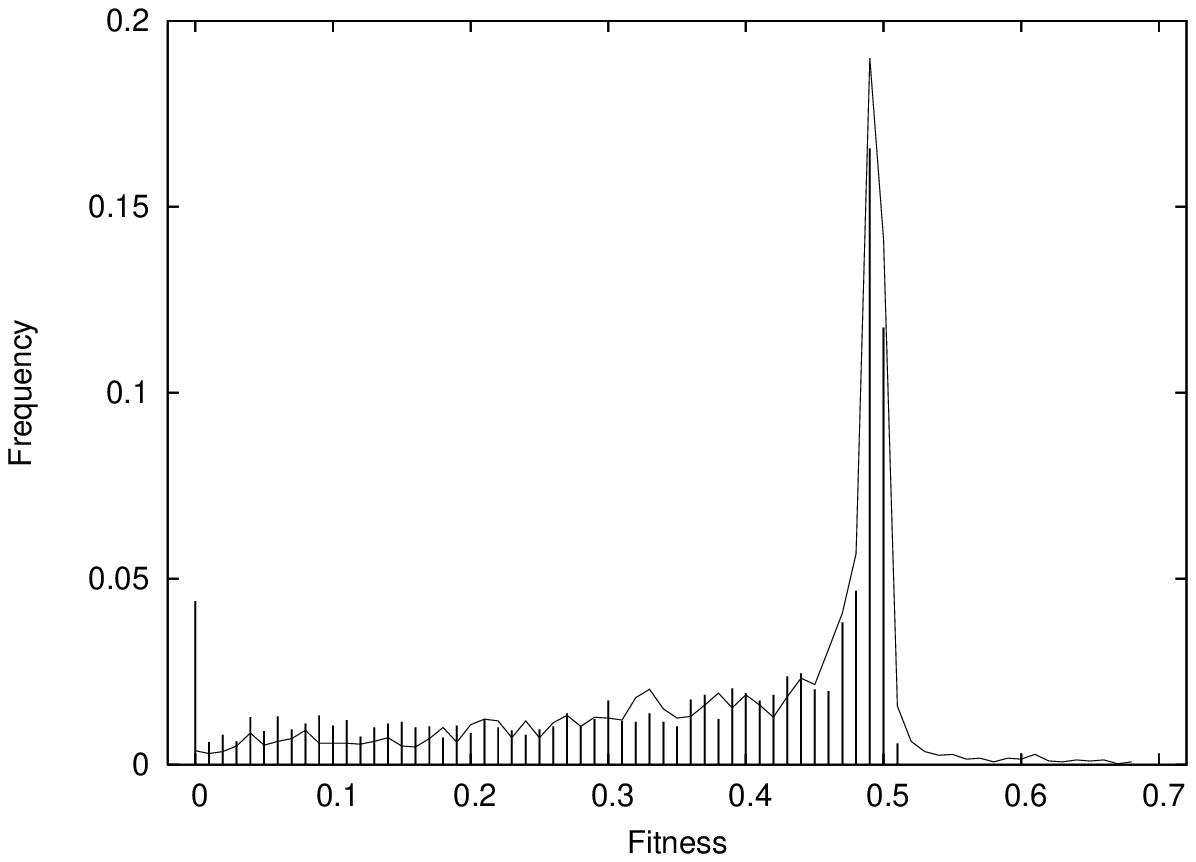}
&
\includegraphics[height=5cm,width=6cm]{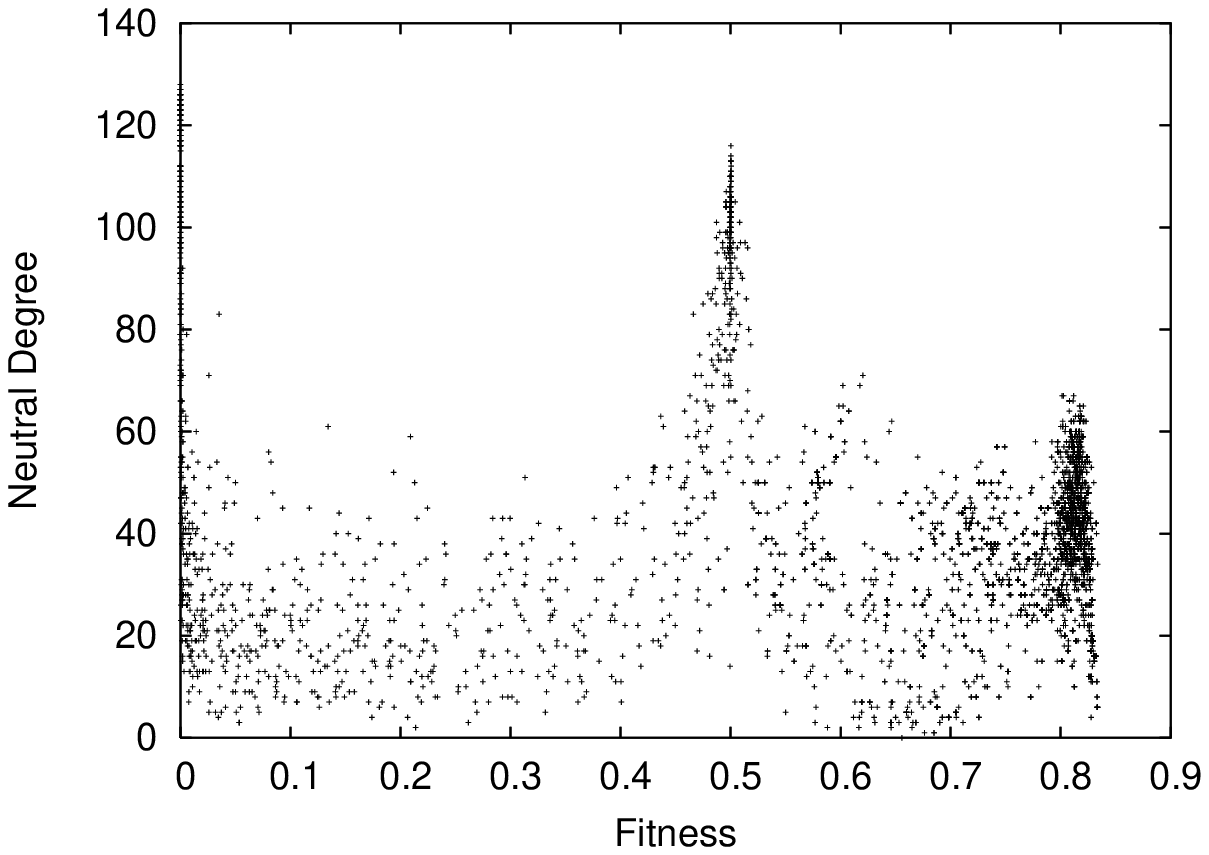}   \\
(a) & (b) \\
\end{tabular}
\end{center}
\caption{(a) DOS using Metropolis-Hasting technique to sample the whole space (impulse) and the Olympus Landscape (line).
(b) Neutral degree on Olympus as a function of the performance. \label{fig-neutr} 
}
\end{figure}

The DOS is more favorable in the Olympus with respect to the whole search space
by sampling the space uniformly at random,
only $28.6 \%$ solutions have null fitness in the random sample. 
Figure \ref{fig-neutr}-a shows the DOS on the Olympus which has been obtained 
by sampling with the Metropolis-Hastings method.
Only $0.3 \%$ solutions have null fitness value in this sample,
although the tail of the distribution is fast-decaying beyond fitness value $0.5$
the highest solution for M-H is $0.68$.
The DOS thus justifies the favours to concentrate the search in the Olympus landscape. 

\paragraph{Neutral Degree.}

The figure \ref{fig-neutr}-b gives the neutral degree of solutions from Olympus 
as a function of their performance.
The solutions below performance $0.5$ are randomly chosen in Olympus.
The solutions over performance $0.5$ are sampled with 2 runs of a GA during $10^3$ generations.
This GA is based on GA defined by Mitchell \cite{mitchelletal94a} 
where the operators are restricted to Olympus subspace
and the selection is a tournament selection of size 2 taking into account the neutrality.
This GA allows to discover a lot of solutions between $0.80$ and $0.835$
and justified the useful of Olympus\footnote{Over $50$ runs, average performances are $0.832$ with standard deviation $0.006$  
which is higher than $0.80_{0.02}$ of coevolutionary algorithm of Pagie \cite{pagie01comparison}}.
Two important $NN$ are located around fitnesses $0$ and $0.5$ where
the neutral degree is over $70$. 
For solutions over $0.5$,
the average of neutral degree is 37.6 which is a high neutral degree.

\section{Discussion and Conclusion} 

The landscape has a considerable number of points with
performance $0$ or $0.5$ which means that
investigations based on sampling techniques on the whole landscape 
are unlikely to give good results. 
The neutrality of the landscape is high, 
and the neutral network topology is not completely random.
Exploiting similarities between the six best rules and symmetries in the landscape, 
we have defined the \textit{Olympus} landscape as a subspace of the Majority problem landscape.
This subspace have less solutions with performance $0$ 
and it is easy to find solutions over $0.80$ with a simple GA.
We have shown that the neutrality of landscape is high 
even for solution over $0.5$.

\bibliographystyle{splncs}

\begin{thebibliography}{10}

\bibitem{wolfram-book-02}
Wolfram, S.:
\newblock A New Kind of Science.
\newblock Wolfram Media (2002)

\bibitem{landbelew95}
Land, M., Belew, R.K.:
\newblock No perfect two-state cellular automata for density classification
  exists.
\newblock Physical Review Letters \textbf{74} (1995)  5148--5150

\bibitem{mitchelletal94a}
Mitchell, M., Crutchfield, J.P., Hraber, P.T.:
\newblock Evolving cellular automata to perform computations: Mechanisms and
  impediments.
\newblock Physica D \textbf{75} (1994)  361--391

\bibitem{andreetal96b}
Andre, D., {Bennett III}, F.H., Koza, J.R.:
\newblock Discovery by genetic programming of a cellular automata rule that is
  better than any known rule for the majority classification problem.
\newblock In Koza, J.R., Goldberg, D.E., Fogel, D.B., Riolo, R.L., eds.:
  Genetic Programming 1996: Proceedings of the First Annual Conference,
  Cambridge, MA, The MIT Press (1996)  3--11

\bibitem{juille98}
Juill\'e, H., Pollack, J.B.:
\newblock Coevolutionary learning: a case study.
\newblock In: ICML '98 Proceedings of the Fifteenth International Conference on
  Machine Learning, San Francisco, CA, Morgan Kaufmann (1998)  251--259

\bibitem{crutch-mitch-das-03}
Crutchfield, J.P., Mitchell, M., Das, R.:
\newblock Evolutionary design of collective computation in cellular automata.
\newblock In Crutchfield, J.P., Schuster, P., eds.: Evolutionary Dynamics:
  Exploring the Interplay of Selection, Accident, Neutrality, and Function.
\newblock Oxford University Press, Oxford, UK (2003)  361--411

\bibitem{packard88}
Packard, N.H.:
\newblock Adaptation toward the edge of chaos.
\newblock In Kelso, J.A.S., Mandell, A.J., Shlesinger, M.F., eds.: Dynamic
  Patterns in Complex Systems.
\newblock World Scientific, Singapore (1988)  293--301

\bibitem{pagie01comparison}
Pagie, L., Mitchell, M.:
\newblock A comparison of evolutionary and coevolutionary search.
\newblock In Belew, R.K., Juill\`{e}, H., eds.: Coevolution: Turning Adaptive
  Algorithms upon Themselves, San Francisco, California, USA (2001)  20--25

\bibitem{dasetal94}
Das, R., Mitchell, M., Crutchfield, J.P.:
\newblock A genetic algorithm discovers particle-based computation in cellular
  automata.
\newblock In Davidor, Y., Schwefel, H.P., {M\"{a}nner}, R., eds.: {PPSN III}.
  Volume 866 of LNCS., Springer-Verlag (1994)  344--353

\bibitem{hansonCrutch95}
Hanson, J.E., Crutchfield, J.P.:
\newblock Computational mechanics of cellular automata: An example.
\newblock Technical Report 95-10-95, Santa Fe Institute Working Paper (1995)

\bibitem{gacsEtal78}
Gacs, P., Kurdyumov, G.L., Levin, L.A.:
\newblock One-dimensional uniform arrays that wash out finite islands.
\newblock Problemy Peredachi Informatsii \textbf{14} (1978)  92--98

\bibitem{reidys01neutrality}
Reidys, C.M., Stadler, P.F.:
\newblock Neutrality in fitness landscapes.
\newblock Applied Mathematics and Computation \textbf{117} (2001)  321--350

\bibitem{KIM:83}
Kimura, M.:
\newblock The Neutral Theory of Molecular Evolution.
\newblock Cambridge University Press, Cambridge, UK (1983)

\bibitem{rose96}
Ros\'e, H., Ebeling, W., Asselmeyer, T.:
\newblock The density of states - a measure of the difficulty of optimisation
  problems.
\newblock In: Parallel Problem Solving from Nature. (1996)  208--217

\bibitem{NIM:99}
Van~Nimwegen, E., Crutchfield, J., Huynen, M.:
\newblock Neutral evolution of mutational robustness.
\newblock In: Proc. Nat. Acad. Sci. USA 96. (1999)  9716--9720

\bibitem{bastolla03}
Bastolla, U., Porto, M., Roman, H.E., Vendruscolo, M.:
\newblock Statiscal properties of neutral evolution.
\newblock Journal Molecular Evolution \textbf{57} (2003)  103--119

\bibitem{gecco-04}
Vanneschi, L., Clergue, M., Collard, P., Tomassini, M., Verel, S.:
\newblock Fitness clouds and problem hardness in genetic programming.
\newblock In: Proceedings of {GECCO}'04. LNCS, Springer-Verlag (2004)

\bibitem{barnett01netcrawling}
Barnett, L.:
\newblock Netcrawling - optimal evolutionary search with neutral networks.
\newblock In: Proceedings of the 2001 Congress on Evolutionary Computation
  CEC2001, COEX, World Trade Center, 159 Samseong-dong, Gangnam-gu, Seoul,
  Korea, IEEE Press (2001)  30--37

\end{thebibliography}

\end{document}